\documentclass[twocolumn]{article}
\usepackage[utf8]{inputenc}
\usepackage{amsmath}
\usepackage{amssymb}
\usepackage{graphicx}
\usepackage{authblk}
\usepackage{tabularx}
\usepackage{booktabs}
\usepackage{hyperref}

\title{Selective Neuron Amplification: A Baseline-Driven Framework for Parameter-Free Task Enhancement in Transformer Models}
\author{Ryyan Akhtar}
\affil{ryyan9ai-ds22@bpitindia.edu.in}
\affil{Dr. Payal Pahwa}
\affil{principal@bpitindia.com}
\affil{Dr. Monika Arora}
\affil{monikaarora@bpitindia.com}
\affil{Department of Artificial Intelligence and Data Science}
\affil{Guru Gobind Singh Indraprastha University, BPIT, New Delhi, India}
\date{} % Hide date

\begin{document}
\maketitle
\thispagestyle{empty}

\begin{abstract}
Large language models often fail at tasks they seem capable of. A model that writes fluent text can still get basic arithmetic wrong, not because the knowledge is missing, but because the circuits responsible for that computation are not consistently activated at inference time. Fine-tuning is the standard fix, but it needs labeled data, extra compute, and risks breaking things that already work. This raises a simpler question: can performance be improved without changing the model at all?
Selective Neuron Amplification (SNA) is our attempt at that. We use differential activation analysis to find neurons that respond more strongly to task-specific inputs, then scale their outputs during a single forward pass using hook-based intervention. The intervention applies only within a single forward pass and leaves model weights unchanged.
We ran 24,192 experiments across 12 tasks covering mathematics, poetry, coding, and logical reasoning on GPT-2 Small and GPT-2 Medium. The main finding was that baseline model confidence predicts SNA effectiveness better than task type does. The relationship is monotonically negative with a Spearman correlation of -0.762 (p = 0.004). We organize this into the Three-Zone Saturation Model. Zone 1 tasks, where baseline probability falls below 0.07, show a mean improvement of 27.85\%, with some configurations reaching 70.3\%. Zone 3 tasks, above 0.10, rarely go beyond 7\% gain. Preliminary results on Pythia-160M show Zone 1 mean improvement of 178.98\% on arithmetic tasks, consistent with the saturation hypothesis. On SST-2 sentiment classification, uncertain predictions improve by 57.3 percentage points while confident ones stay stable.
The practical takeaway is straightforward: measure baseline confidence first, and apply SNA only where the model is under-confident.
\end{abstract}

\section{Introduction}
\subsection{Motivation \& The Approach}

A model that writes fluent text can still get basic arithmetic wrong, not because the knowledge is missing, but because the circuits responsible for that computation are not consistently activated at inference time. The capability is there, it just does not show up when it needs to.

The standard response is fine-tuning. Train the model on task-specific data, update the weights, and expect performance to improve. This works in many cases, but it is often more intervention than the problem requires. Fine-tuning needs labeled data and real compute. Models can also lose capabilities they already had, and updates sometimes just teach the model to better match expected output formats rather than actually improving reasoning.

If the real issue is underactivated circuits, the question becomes whether you can just turn up the signal instead of retraining the whole model, and that is the central idea behind Selective Neuron Amplification, and it motivates this work.

Differential activation analysis identifies neurons that respond more strongly to task-specific inputs compared to neutral reference text. Using TransformerLens hooks \cite{nanda2022transformerlens}, their outputs are scaled during a single forward pass by a factor between 1.5x and 3.0x. Nothing in the model is permanently changed.
Prior work has explored inference-time steering in different ways. CAA \cite{panickssery2024caa} builds steering vectors from contrastive examples. ACTIVSCALAR \cite{stoehr2024activscalar} learns multiplicative scaling factors through gradient-based optimization. Both can shift model behavior without touching weights, but neither tells you when steering is actually worth trying. Stolfo et al. (2024) showed that output confidence is actively regulated by identifiable neurons, providing mechanistic grounding for why baseline probability is a useful pre-intervention signal.

The central claim of this work is that baseline model confidence, measured before any intervention, reliably predicts whether amplification will help. Low-confidence outputs tend to improve. High-confidence outputs mostly do not.
An extended version of this work, including full experimental logs and the SNA tool documentation, is available as a preprint \cite{akhtar2026sna}.

\subsection{Contributions}
We make the following contributions in this paper.

\begin{enumerate}
    \item The Three-Zone Saturation Model.
    
    Baseline confidence predicts SNA outcomes better than task type or difficulty. Zone 1 (baseline below 0.07) yields mean improvement of 27.85\%, reaching 70.3\% in favorable configurations. Zone 2 (0.07 to 0.10) is unstable and sensitive to circuit localization. Zone 3 (above 0.10) rarely exceeds 7\%, a ceiling that appears structural.
    \item A Predictive Three-Factor Framework. 
    
    Baseline capability accounts for roughly 58\% of variance in rank-ordered outcomes, circuit localization for 28\%, and task type for the remaining 14\%.
    \item Functional Layer Specialization.
    
    Layer 8 recurs as the optimal intervention point for arithmetic, Layer 10 for logical reasoning, Layer 21 for abstract semantic tasks, and Layer 22 for syntactic pattern matching in code.
    \item Cross-Task Interference.
    
    Amplifying arithmetic circuits at Layer 8 suppresses poetry performance by 8 to 12 percentage points. The reverse does not hold, indicating circuits compete for shared representational capacity rather than operating independently.
    \item Generalization to Classification.
    
    SNA applied to SST-2 sentiment classification improves Zone 1 predictions by up to 59.1 percentage points while Zone 3 predictions show zero change across all 60 tested configurations.
\end{enumerate}

\section{Background \& Related Work}
\subsection{Transformer FFN Neurons}

Each transformer layer has two main components: a multi-head attention mechanism and a feed-forward network. Attention handles relationships between tokens. The feed-forward network operates on each token independently, applying learned transformations before passing the result to the next layer. SNA only touches the feed-forward component.
Geva et al. (2021) \cite{geva2021transformer} showed that feed-forward neurons function as key-value memories, developing specialized responses during training that align with identifiable input patterns. In GPT-2 Medium, each feed-forward network contains 4,096 neurons. Some become sensitive to arithmetic structures, others to poetic form, others to logical relationships. This specialization emerges naturally from training as the model learns which internal features are most useful for predicting the next token in different contexts.
That structure is what makes targeted intervention possible. If neurons responded uniformly to all input types, there would be no meaningful target to amplify.

\subsection{Related Work}

Several methods have explored shaping model behavior at inference time without updating weights.
\begin{enumerate}
    \item Turner et al. (2023)\cite{turner2023activation} showed this was possible by adding difference vectors to GPT-2-XL's residual stream to redirect outputs. Zou et al. (2023) extended this idea using representation engineering, working with directions in activation space to target higher-level concepts. Olah et al. (2020) and Wang et al. (2023) provided earlier mechanistic groundwork showing that transformer circuits develop identifiable, interpretable structure during training. SNA builds on this intuition but does not inject anything external. It only amplifies signals already present in the network.
    \item Panickssery et al. (2024) \cite{panickssery2024caa} introduced CAA, which builds steering vectors from contrastive behavioral examples and injects them at inference time, showing real effects on Llama 2 for sycophancy and hallucination. Layer choice and scaling strength are figured out after running sweeps, making it reactive. SNA measures baseline confidence before any intervention and uses that to decide whether amplification is worth trying at all.
    \item ACTIVSCALAR (Stoehr et al., 2024) \cite{stoehr2024activscalar} is closest to SNA in spirit. It learns multiplicative scaling factors through gradient-based optimization to correct specific errors, but requires an optimization step and training signal. SNA is zero-shot and parameter-free.
    \item Stolfo et al. (2024) \cite{stolfo2024confidence} found neurons in GPT-2 that regulate output confidence through LayerNorm scaling without changing token rankings. This helps explain why baseline probability tracks SNA responsiveness. Low confidence likely reflects underactivated circuits, which is exactly when amplification tends to work.
\end{enumerate}

\subsection{Why GPT-2?}

GPT-2 Small (117M parameters) and GPT-2 Medium (345M parameters) were chosen for three reasons. First, their capability gaps are visible and measurable. Basic arithmetic works sometimes, multi-step calculations mostly do not. Any improvement from SNA can be reasonably attributed to the intervention rather than pre-existing model strength. More capable models complicate this because the signal gets buried under general competence. 

Second, both models are open-source, well-documented, and runnable on consumer hardware. Running over 24,000 experiments on a closed or proprietary system would have been a real constraint. 

Third, GPT-2's well-documented capability gaps are analytically useful here. SNA needs tasks where the model actually struggles, and these models provide exactly that. They are best understood as test beds for properties like circuit saturation, layer specialization, and confidence-dependent behavior, all of which likely generalize to larger transformer architectures.

\section{Methodology}
\subsection{SNA Mechanics}
\textbf{Neuron Identification-}
The model is run on two sets of inputs: task-specific examples like arithmetic problems, and neutral reference text like simple descriptive sentences. For each neuron in every layer, we compute the difference in mean activation between these two conditions. Neurons with the largest positive differences are responding to the task rather than to text in general. These are selected as candidates for amplification.

\textbf{Intervention-}
We register a hook at the feed-forward output of the chosen layer using TransformerLens. During the forward pass, this hook intercepts activations, scales selected neurons by a factor between 1.5x and 3.0x, and passes modified values back into the network. No parameters are changed. Once the hook is removed, the model returns exactly to its original state. This is what made running 24,000 plus experiments on the same base model practical.

\textbf{Why This Is More Than Simple Scaling-}
The selected neurons project into the residual stream, so scaling their outputs has downstream consequences across all subsequent layers. The modification is local in scope but non-local in effect. No new information enters the model. SNA only amplifies representations already present but contributing too weakly to influence predictions under normal inference. This has a direct implication: if the relevant representation does not exist, scaling cannot recover it. That is the reason measuring baseline confidence before applying SNA matters.

\subsection{Task Selection}
The core question was whether baseline confidence predicts SNA effectiveness or whether task type matters more. To test this, we needed tasks that naturally produce a spread of baseline confidence values both within and across domains. Four domains were chosen with this in mind.

Mathematics was the clearest choice. GPT-2 sees very little arithmetic during pretraining, so baseline probabilities tend to be low across all three difficulty levels, giving us a reliable source of Zone 1 conditions.

Poetry works differently. Confidence drops not because the task is computationally hard, but because the prompt demands abstraction. Simple rhyming tasks produce fairly confident outputs while metaphorical prompts do not, letting us observe zone transitions within a single domain.

Coding was included as a deliberate contrast. GPT-2 was trained on large amounts of code, so we expected high baselines across all difficulty levels. Limited SNA response here would reflect circuit saturation, not task difficulty.
Logic was harder to place. GPT-2 has no formal logic training but responds confidently to logical patterns, likely from surface-level pattern matching over familiar structures in pretraining data. Saturation should still appear, just for a different reason than in coding.

Across four domains and three difficulty levels each, we had twelve tasks total, designed so baseline confidence and task type could be at least partially separated in the analysis.

\begin{table}[htbp]
    \centering
    \includegraphics[width=80mm]{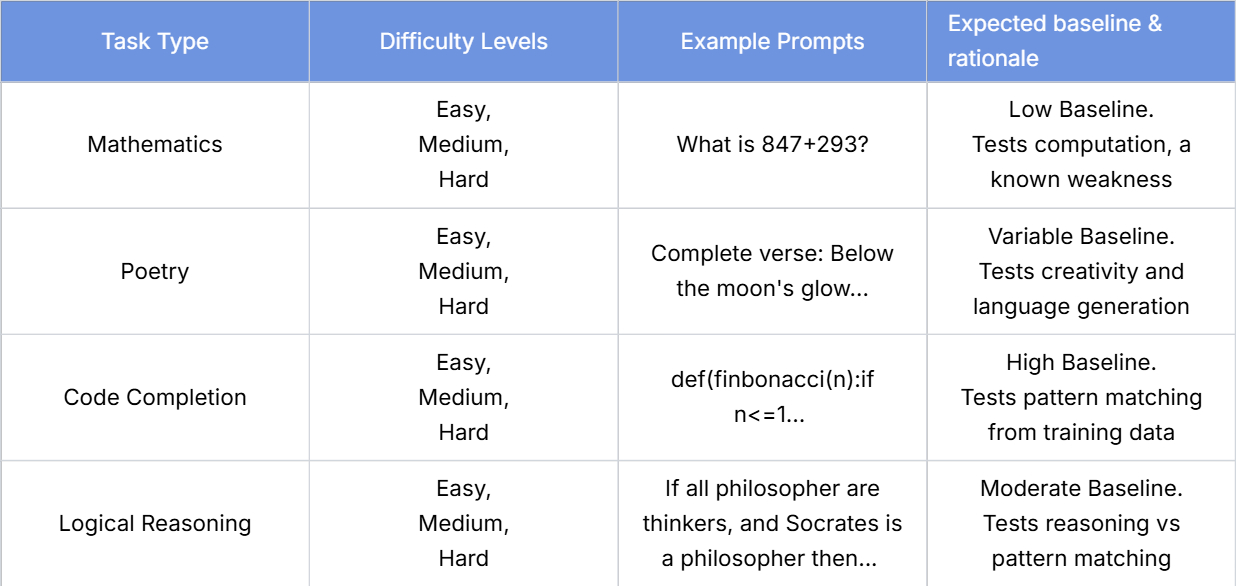}
    \caption{Task Overview}
    \label{fig:waveform_classical}
\end{table}

\subsection{Parameter Sweep}
For each of the 12 tasks, we varied three parameters in a full factorial design. Layer selection covered all 24 layers in GPT-2 Medium (0 to 23), tested exhaustively rather than selected based on prior assumptions. 

This lets the specialization pattern emerge from the data. Neuron count was varied across 7 values: 3, 5, 8, 10, 15, 20, and 25. Smaller counts target localized circuits with more precision while larger counts cast a wider net for distributed processing. Amplification multiplier was varied across 12 values from 1.1x to 2.4x, spanning from negligible effect to the upper bound before destabilization risk increases.

This gives 2,016 unique configurations per task (24 layers times 7 neuron counts times 12 multipliers) and 24,192 total experiments across all 12 tasks. Running all combinations rather than a hand-selected subset is what allows us to make claims about consistency and reproducibility across the full parameter space. Cherry-picked configurations can look good without being representative.

\subsection{Evaluation Metrics}
We evaluate SNA using four metrics applied consistently across all experiments.

\begin{enumerate}
    \item Baseline probability is the model's confidence in the correct output before any intervention, expressed as the probability assigned to the target token. This is the primary independent variable used to interpret SNA outcomes.
    \item Percentage improvement quantifies the effect of amplification for each configuration, calculated as:
    \[
    \left( \frac{P_\text{post} - P_\text{base}}{P_\text{base}} \right) \times 100\%
    \] 
    Normalizing by baseline allows fair comparison across tasks with very different starting confidence levels.
    \item Success rate measures the proportion of configurations that produce any positive improvement for a given task. This captures whether SNA generally helps or only works under specific parameter settings.
    \item Golden zone count identifies configurations that achieve more than 10\% improvement. This is particularly important because it shows how broadly SNA works across the parameter space, not just what the single best run managed to produce.
\end{enumerate}

\section{Results}

\begin{figure}
    \centering
    \includegraphics[width=1\linewidth]{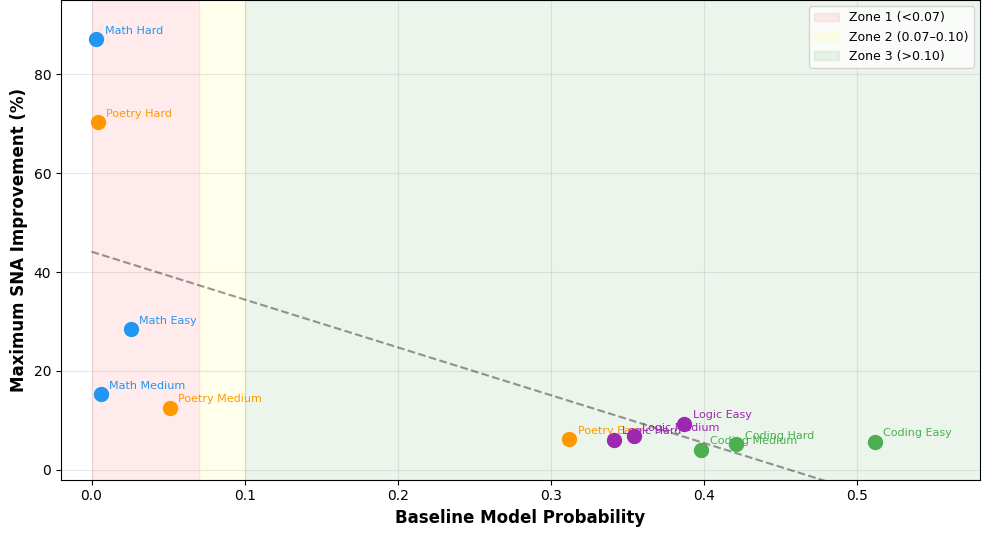}
    \caption{Inverse relationship between baseline confidence and SNA improvement across tasks.}
    \label{fig:placeholder}
\end{figure}

\subsection{Baseline as the Governing Variable}
Across all 12 tasks, the relationship between baseline confidence and SNA improvement is consistently and robustly negative.
Pearson gives r = -0.503 (p = 0.095, R² = 0.253), a moderate negative trend that falls short of significance. With only 12 tasks, linear correlation is not a reliable measure here.
Spearman is clearer: $\rho = -0.762$ ($p = 0.004$), a strong monotonic relationship. 
$\rho^2 = 0.581$ means baseline rank accounts for roughly 58\% of variance in rank-ordered improvement. 

A bootstrap test across 10{,}000 resamples gives a 95\% confidence interval of $-0.918$ to $-0.351$. 
It stays negative throughout, with no single task driving the trend.
Tasks below 0.07 regularly show large gains, sometimes reaching 70\%. Above 0.10, improvements shrink to single digits. Zone 1 averages 27.85\%. Zone 3 stays below 8\%.

This is where SNA differs from CAA \cite{panickssery2024caa} and ACTIVSCALAR \cite{stoehr2024activscalar}. Both decide how to intervene after running sweeps. SNA uses baseline confidence to decide whether to intervene at all, before touching anything. That idea is what the Three-Zone Saturation Model formalizes.

\subsection{Zone 1 — Optimal Responsiveness (Baseline $< 0.07$)}

Five tasks fell into Zone 1: all three mathematics levels, poetry medium, and poetry hard. Mean improvement was 27.85\%, with the peak reaching 70.3\%.
Mathematics is the cleanest case. Baselines sit between 0.003 and 0.026, reflecting how little arithmetic GPT-2 saw during training. Those numbers might suggest the capability is absent, but the SNA results say otherwise. Something is present in the network, just not contributing strongly enough under normal inference. 

Layer 8 appears as the optimal intervention point across all three math difficulty levels. Difficulty does not shift the optimal layer, only domain does.
Poetry is messier. Layer 21 comes up most often, but results vary considerably across prompts. Abstract language draws on more distributed semantic representations, which are harder to target cleanly than the localized arithmetic circuits at Layer 8.

\subsection{Zone 3 — Saturation Ceiling (Baseline $> 0.10$)}

Seven tasks landed in Zone 3: all coding levels, all logic levels, and poetry easy. We ran 2,016 configurations per task and still saw minimal improvement. At that point the issue is clearly structural, not a tuning problem.
Coding makes sense. GPT-2 saw a lot of code during pretraining, and baselines between 0.32 and 0.51 reflect that. The circuits are already fully engaged, there is nothing left to amplify.

Logic is harder to read. The high confidence likely comes from pattern matching over familiar structures rather than genuine reasoning. But from SNA's perspective that distinction does not change anything. Saturated circuits do not respond to amplification regardless of why they got saturated. This does expose a real limitation though: the method cannot tell the difference between genuine capability and learned surface patterns. Both look the same from the outside.

\subsection{Zone 2 — Transition Region (0.07–0.10)}

Almost no tasks fell in this range, and that was not intentional. It might say something about how GPT-2 organizes capabilities internally. Skills seem to land at one of two extremes: either well-reinforced with high confidence, or weakly represented with low confidence. There is not much in between. Behavior in this range is expected to be unstable and sensitive to parameter choices. Testing Zone 2 properly would require tasks deliberately constructed to hit the 0.07 to 0.10 baseline range, which remains an open direction.

\subsection{Predictive Three-Factor Framework}
Baseline confidence is the strongest predictor of SNA outcomes, accounting for roughly 58\% of variance in rank-ordered results ($\rho^2 = 0.581$). Circuit localization contributes an estimated 28\%. Arithmetic maps consistently to Layer 8, which makes its gains more predictable. Poetry is less stable across prompts because abstract language draws on more distributed representations at Layer 21. Task type explains the remaining 14\%. Computational tasks with similar baselines tend to show larger gains than semantic ones.

These proportions are working approximations, not fixed constants. But they give a practical ordering for deciding whether amplification is worth attempting: check baseline confidence first, then assess how localized the relevant circuits are, then consider task type.

\section{Architectural Analysis}

\subsection{Layer Specialization}
Layer specialization was a secondary finding. The experiments were designed to test when SNA works, not where. The pattern emerged from the results: peak performance kept appearing at the same layer depths, varying by domain rather than difficulty. 
Mathematics peaked at Layer 8 across all three difficulty levels, poetry at Layer 21, logic at Layer 10, coding at Layer 22. Difficulty did not shift the optimal layer within any domain. All peaks are averaged across neuron counts and multiplier values, not single best-configuration results.

\begin{itemize}
    \item Layer 8 — Arithmetic.
    Neurons here respond strongly to numerical inputs compared to ordinary text. The same intervention applied to poetry or coding at Layer 8 produces almost no effect, which confirms the specificity.

    \item Layer 10 — Logical Operations.
    Neurons respond selectively to inference-related tokens like "if," "then," and "therefore." Average gain is 6.8\%, expected for Zone 3, but localization is stable across configurations.
    
    \item Layer 21 — Abstract Semantic Processing.
    Largest improvements across all layers, and optimal for both poetry and SST-2 sentiment classification. Both tasks rely on high-level semantic content like emotional tone and figurative meaning. Neurons here respond to full-sentence context rather than individual tokens, consistent with the depth at which this layer sits.

    \item Layer 22 — Syntactic Pattern Matching.
    Coding peaked here, the deepest of the four. Code completion is surface pattern recognition, and token-level patterns are most resolved near the output end. Average improvement was 3.2\%, consistent with Zone 3 saturation.
\end{itemize}
\subsection{Cross-Task Interference}

To check whether specialized layers operate independently, we ran cross-task interference experiments. Applying the Layer 8 arithmetic configuration (15 neurons, 2.5x) and measuring its effect on poetry generation in the same forward pass dropped poetry fluency by 8 to 12 percentage points. Reversing the experiment, applying Layer 21 poetry surgery and measuring arithmetic performance, showed no meaningful change.
The asymmetry follows from how the network processes sequentially. Layer 8 sits early in the pipeline, so changes there propagate forward through all subsequent computation including Layer 21. Layer 21 sits near the output end, so arithmetic computation at Layer 8 is already complete before any modification happens there.
These circuits share the residual stream, and amplifying one reshapes the representational context for everything computed downstream. A clean modular picture where each circuit operates independently does not hold here. Targeting one capability can produce side effects on others, and the direction of those effects depends on where in the network the intervention is applied.

\section{Validation: SST-2 Sentiment Classification}
\subsection{Motivation \& Setup}

Controlled generation tasks isolate variables well, but the framework needs to hold on something closer to a real use case. Sentiment analysis maps naturally onto the zone structure: obvious reviews should sit in Zone 3 where circuits are already well-exercised, and ambiguous ones should land in Zone 1 where those circuits are less active. SST-2 from the GLUE benchmark was a practical choice given its reliable labels and wide use.

GPT-2 Medium was tested zero-shot with no fine-tuning. Each example was formatted as "Review: [sentence] Sentiment:" and predictions came from comparing probabilities for " positive" versus " negative." We used 200 examples from the SST-2 validation set with seed 42.

Applying the original zone thresholds directly did not work. GPT-2 operates over a 50,257-token vocabulary, so absolute token probabilities are almost always very low, and every example fell into Zone 1 regardless of how confidently the model separated the two labels. We defined a confidence margin instead:
\[
\text{Margin} = \frac{\left| P(\text{positive}) - P(\text{negative}) \right|}{P(\text{positive}) + P(\text{negative})}
\]
Zone assignments were: Zone 1 (margin below 0.30, n=66), Zone 2 (0.30 to 0.60, n=89), Zone 3 (above 0.60, n=45). Baseline accuracy before any intervention was 34.8\% in Zone 1, 75.3\% in Zone 2, and 84.4\% in Zone 3, which validated the adaptation.
Standard differential activation against neutral text gave 53\% overlap between neurons flagged for positive and negative sentiment, making targeted amplification unusable. Switching to direct contrastive analysis, comparing positive examples directly against negative ones, resolved this. At Layer 21, we identified 10 neurons exclusively tied to positive sentiment and 10 exclusively tied to negative, with zero overlap.

\subsection{Results}

\begin{figure}
    \centering
    \includegraphics[width=1\linewidth]{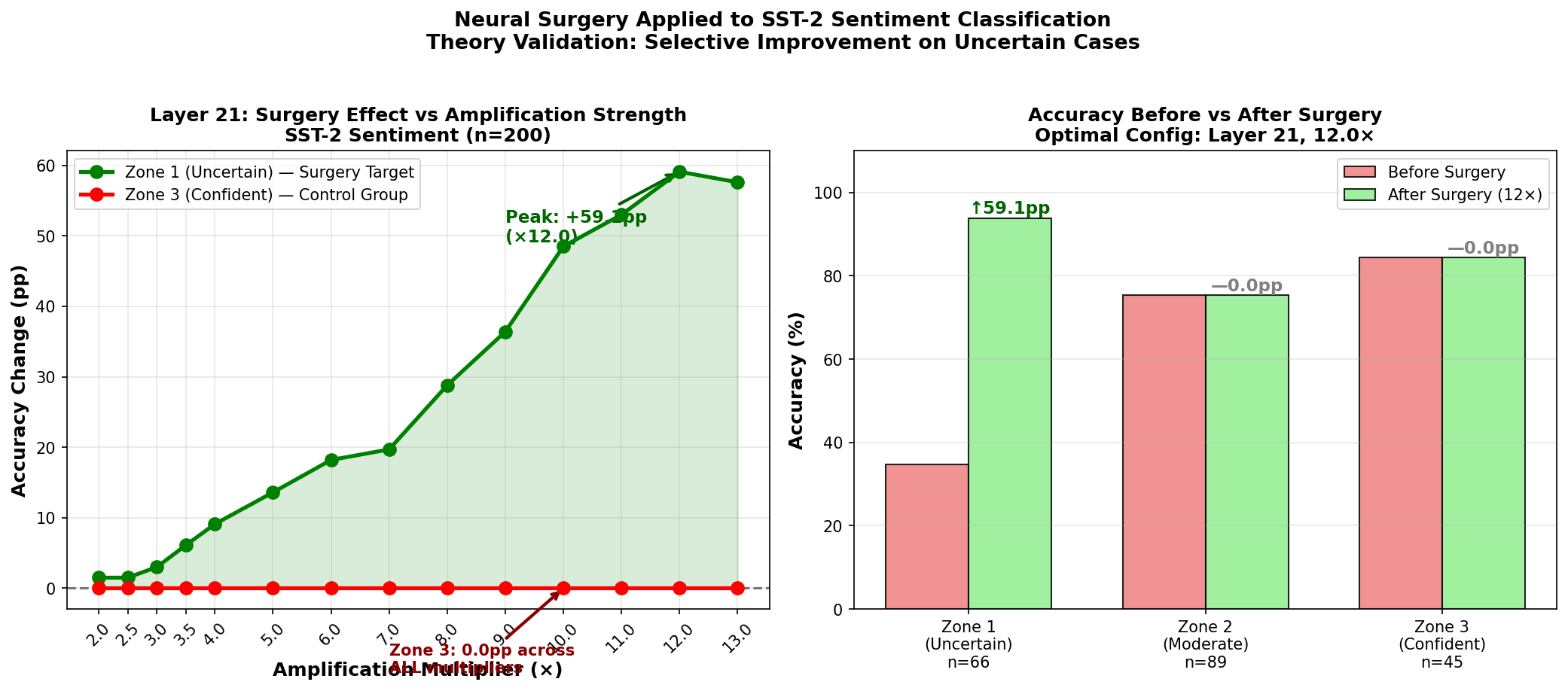}
    \caption{SST-2 sentiment classification results at Layer 21}
    \label{fig:placeholder}
\end{figure}

Layer 21 was the only layer that produced consistent Zone 1 improvement across the grid search (5 layers, 4 multipliers). Every other layer either had no effect or actively hurt performance. This fits with what Section 5 found about Layer 21's role in abstract semantic processing.
The split between Zone 1 and Zone 3 is about as clean as anything we observed in this study. Zone 1 gains reach up to 59.1 percentage points at higher multipliers. Zone 3 sits at exactly 0.0pp across all 60 tested configurations, no exceptions. The Zone 1 response curve starts flat at low multipliers, climbs steeply, then plateaus. That shape is more consistent with crossing a decision boundary than a smooth probability increase. Zone 3 does not move at any multiplier.
The multipliers needed here peak around 12x, much higher than the 2 to 3x range that worked in generation tasks. The reason is structural. In generation, the model just needs to assign higher probability to the correct token. In classification, the correct label has to beat out an alternative. The bar is higher, but the underlying mechanism is the same.
Four claims from the core framework are confirmed here. Baseline confidence predicts effectiveness. Zone 3 saturation is structural. Layer 21 specialization generalizes beyond generation. The zone model applies to discriminative tasks.

\section{Discussion}
\subsection{Computation vs. Retrieval vs. Creation}

Different tasks respond to SNA in different ways, and the pattern tracks how each task is represented internally.

Arithmetic is the clearest case. Baselines are near zero yet gains are unusually consistent across configurations. If the capability were simply absent, amplification would produce erratic results. The consistency suggests the circuits are present, just not contributing strongly enough under normal inference. SNA works here because it does not need to create anything, only amplify what already exists.

Coding and logic sit in high-confidence territory and neither responds much to amplification. With coding this makes sense given how much code GPT-2 saw during training. Logic is murkier, likely pattern matching over familiar structures rather than genuine reasoning. From SNA's perspective this distinction does not change the outcome. A circuit saturated through pattern retrieval behaves the same as one saturated through real competence, and the method has no way to tell them apart.

Poetry is less predictable. The network draws on different internal resources depending on what the prompt demands, making gains harder to pin down and the optimal configuration less stable than arithmetic.
Classification introduces threshold-crossing behavior. Performance stays flat at low multipliers then improves rapidly once the decision boundary is crossed, which is why amplification factors around 12x are needed compared to 2 to 3x in generation tasks.

CAA and SNA are sometimes described in similar terms but operate under different assumptions. CAA shifts representations in a chosen direction, useful when the model is producing the wrong behavior. SNA assumes the direction is already roughly correct but under-expressed. They are not interchangeable.

\subsection{The SNA Tool}

To make the framework accessible, we built an interactive demonstration tool using Streamlit, hosted publicly on Hugging Face Spaces at \href{https://huggingface.co/spaces/ry06/Selective_Neuron_Amplification}{Selective Neuron Amplification Demo} The tool supports five models: GPT-2 Small, Medium, Large, XL, and Pythia-160M. Users can measure baseline confidence, get zone predictions, and apply SNA without writing any code. This is a research demonstration, not a production system, and inference times will vary depending on model size and available hardware.
Full experimental details and additional results are available in the preprint version of this work \cite{akhtar2026sna}

\subsection{Generalization}

The core relationship between training frequency and circuit strength is a general property of gradient-based learning, not something specific to GPT-2. Circuits that are exercised repeatedly during pretraining produce stronger, more consistent activations. Ones that are rarely activated stay weak. That imbalance likely holds across transformer architectures.
What does not transfer directly are the specific details. The 0.07 threshold, optimal layer indices, and zone boundaries are tied to GPT-2 Medium and the tasks used here. Different architectures and training distributions will shift these values. The framework may generalize, but the parameters need recalibration for each new setting.
Preliminary results on Pythia-160M are consistent with this view. Across Zone 1 arithmetic configurations with baselines between 0.0002 and 0.0032, mean improvement reached 178.98\%. The optimal layer shifted from Layer 8 in GPT-2 Medium to Layer 0 in Pythia-160M, which has 12 layers compared to 24. This suggests intervention depth scales with overall model depth. These results are promising but still limited in scope, and should be read as motivation for cross-architecture investigation rather than confirmation of any structural claim.

\section{Limitations \& Future Work}

\textbf{Limitations}

\begin{enumerate}
    \item Model scope. Core experiments are on GPT-2 Small and Medium. Pythia-160M results are preliminary cross-architecture evidence, not systematic validation. Zone thresholds and optimal intervention layers have not been mapped across other model families.
    \item Zone 2 gap. No tasks fell in the 0.07 to 0.10 baseline range. Transition behavior here is inferred from boundary cases in the poetry domain, not directly tested. Tasks deliberately constructed to fall within this range are needed to validate the framework fully.
    \item High multipliers in classification. Sentiment classification required multipliers around 12x compared to 2 to 3x in generation. Practical effects at this scale, including output instability and interference in longer sequences, have not been tested.
    \item Single-task surgery only. Every experiment amplifies one task at a time. Simultaneous multi-task amplification and its interaction with cross-task interference remains unexplored.
    \item Zero-shot classification baseline. GPT-2 Medium achieved 64\% on SST-2 zero-shot, well below fine-tuned classifiers. How SNA behaves on already fine-tuned or instruction-tuned models is an open question.
\end{enumerate}

\textbf{Future Work}

\begin{enumerate}
    \item The most immediate direction is systematic validation on modern architectures like LLaMA, Mistral, and Gemma, with full zone threshold recalibration per model. The 0.07 boundary and optimal layer indices are specific to GPT-2 Medium and should not be assumed to hold elsewhere.
 
    \item Constructing tasks that deliberately fall in the 0.07 to 0.10 range would allow direct empirical testing of Zone 2 behavior. Transition-region predictions are currently inferred, not tested, which is the weakest part of the framework.
 
    \item The cross-task interference findings suggest multi-task surgery is worth exploring. The asymmetry between Layer 8 and Layer 21 interference indicates that simultaneous interventions at non-overlapping layers might be compatible without degrading either task.
 
    \item Finally, combining SNA with activation patching or causal tracing would move the analysis from layer-level localization to neuron-level interpretability. We know where to intervene. What those neurons are actually computing remains an open question.
\end{enumerate}

\section{Conclusion}
The starting point of this work was a simple observation: language models sometimes fail at tasks they appear to know how to do. Our experiments suggest this gap is often not about missing knowledge. In a number of cases, the relevant circuits are present but insufficiently activated, and scaling their outputs at inference time is enough to shift the outcome. SNA addresses this through a parameter-free, single-pass, fully reversible intervention.

The clearest finding across 24,192 experiments was that baseline model confidence predicts SNA effectiveness better than task type does. The Spearman correlation of negative 0.762 (p = 0.004) holds consistently across domains and model sizes. The Three-Zone Saturation Model organizes this into a practical guide: measure baseline confidence first, and apply SNA selectively in Zone 1 where circuits have room to respond. Validation on SST-2 sentiment classification confirms the zone pattern holds in discriminative settings, with Zone 1 gains of up to 59.1 percentage points and zero change on Zone 3 predictions across all 60 tested configurations. Cross-task interference experiments further show that these circuits are not independent, and that the direction of interference follows from network depth.

These results are validated on GPT-2 Small and Medium. The 0.07 zone boundary and optimal layer indices are specific to these models and will need recalibration for other architectures. The broader implication is that some performance gaps in language models do not require retraining to address. If the capability is present but underexpressed, the right intervention is amplification, not adaptation.

\bibliographystyle{plain}
\nocite{*}
\bibliography{bib}

@article{radford2019language,
  title   = {Language Models are Unsupervised Multitask Learners},
  author  = {Radford, Alec and Wu, Jeffrey and Child, Rewon and
             Luan, David and Amodei, Dario and Sutskever, Ilya},
  journal = {OpenAI Blog},
  volume  = {1},
  number  = {8},
  pages   = {9},
  year    = {2019}
}

@article{stolfo2024confidence,
  title={Confidence Regulation Neurons in Language Models},
  author={Stolfo, Alessandro and Wu, Benjie and Gurnee, Wes and Belinkov, Yonatan and Song, Xingyi and Sachan, Mrinmaya and Nanda, Neel},
  journal={arXiv preprint arXiv:2406.16254},
  year={2024},
  url={https://arxiv.org/abs/2406.16254}
}

@article{olah2020zoom,
  title   = {Zoom In: An Introduction to Circuits},
  author  = {Olah, Chris and Cammarata, Nick and Schubert, Ludwig and
             Goh, Gabriel and Petrov, Michael and Carter, Shan},
  journal = {Distill},
  year    = {2020},
  doi     = {10.23915/distill.00024.001},
  url     = {https://distill.pub/2020/circuits/zoom-in/}
}

@inproceedings{geva2021transformer,
  title     = {Transformer Feed-Forward Layers are Key-Value Memories},
  author    = {Geva, Mor and Schuster, Roei and Berant, Jonathan and
               Levy, Omer},
  booktitle = {Proceedings of the 2021 Conference on Empirical Methods
               in Natural Language Processing},
  pages     = {5484--5495},
  year      = {2021},
  publisher = {Association for Computational Linguistics}
}

@article{zou2023representation,
  title   = {Representation Engineering: A Top-Down Approach to
             {AI} Transparency},
  author  = {Zou, Andy and Phan, Long and Chen, Sarah and Campbell, James and
             Guo, Phillip and Ren, Richard and Pan, Alexander and Yin, Xuwang and
             Mazeika, Mantas and Dombrowski, Ann-Kathrin and Goel, Shashwat and
             Li, Nathaniel and Byun, Michael J and Wang, Zifan and
             Mallen, Alex and Basart, Steven and Koyejo, Sanmi and
             Song, Dawn and Fredrikson, Matt and Kolter, J Zico and
             Hendrycks, Dan},
  journal = {arXiv preprint arXiv:2310.01405},
  year    = {2023}
}

@article{turner2023activation,
  title   = {Activation Addition: Steering Language Models Without
             Optimization},
  author  = {Turner, Alexander Matt and Thiergart, Lisa and Udell, David and
             Leike, David and Mini, Ulisse and MacDiarmid, Monte},
  journal = {arXiv preprint arXiv:2308.10248},
  year    = {2023}
}

@inproceedings{panickssery2024caa,
  title     = {Steering {Llama} 2 via Contrastive Activation Addition},
  author    = {Panickssery, Nina and Gabrieli, Nick and Schulz, Julian and
               Tong, Meg and Hubinger, Evan and Turner, Alexander Matt},
  booktitle = {Proceedings of the 62nd Annual Meeting of the
               Association for Computational Linguistics},
  pages     = {15504--15522},
  year      = {2024},
  publisher = {Association for Computational Linguistics},
  url       = {https://aclanthology.org/2024.acl-long.828}
}

@inproceedings{stoehr2024activscalar,
  title     = {Activation Scaling for Steering and Interpreting
               Language Models},
  author    = {Stoehr, Niklas and Du, Kevin and Sn{\ae}bjarnarson,
               V{\'e}steinn and West, Robert and Cotterell, Ryan and
               Schein, Aaron},
  booktitle = {Findings of the Association for Computational Linguistics:
               EMNLP 2024},
  year      = {2024},
  url       = {https://aclanthology.org/2024.findings-emnlp.479}
}

@article{baroni2025layernorm,
  title={Transformers Don't Need {LayerNorm} at Inference Time: Scaling {LayerNorm} Removal to {GPT-2} {XL} and the Implications for Mechanistic Interpretability},
  author={Baroni, Luca and Khara, Galvin and Schaeffer, Joachim and Subkhankulov, Marat and Heimersheim, Stefan},
  journal={arXiv preprint arXiv:2507.02559},
  year={2025},
  url={https://arxiv.org/abs/2507.02559}
}

@inproceedings{wang2018glue,
  title     = {{GLUE}: A Multi-Task Benchmark and Analysis Platform for
               Natural Language Understanding},
  author    = {Wang, Alex and Singh, Amanpreet and Michael, Julian and
               Hill, Felix and Levy, Omer and Bowman, Samuel R},
  booktitle = {Proceedings of the 2018 EMNLP Workshop BlackboxNLP:
               Analyzing and Interpreting Neural Networks for NLP},
  pages     = {353--355},
  year      = {2018},
  publisher = {Association for Computational Linguistics}
}

@inproceedings{biderman2023pythia,
  title     = {Pythia: A Suite for Analyzing Large Language Models
               Across Training and Scaling},
  author    = {Biderman, Stella and Schoelkopf, Hailey and
               Anthony, Quentin Gregory and Bradley, Herbie and
               O'Brien, Kyle and Hallahan, Eric and Khan, Mohammad Aflah and
               Purohit, Shivanshu and Prashanth, USVSN Sai and Raff, Edward and
               Skowron, Aviya and Sutawika, Lintang and Van Der Wal, Oskar},
  booktitle = {Proceedings of the 40th International Conference on
               Machine Learning},
  pages     = {2397--2430},
  year      = {2023},
  publisher = {PMLR}
}

@software{nanda2022transformerlens,
  title  = {{TransformerLens}},
  author = {Nanda, Neel and Bloom, Joseph},
  year   = {2022},
  url    = {https://github.com/neelnanda-io/TransformerLens},
  note   = {A Library for Mechanistic Interpretability of
            GPT-Style Language Models}
}

@software{streamlit2023,
  title  = {Streamlit: The Fastest Way to Build and Share Data Apps},
  author = {{Streamlit Inc.}},
  year   = {2023},
  url    = {https://streamlit.io}
}

@misc{akhtar2026sna,
  author       = {Ryyan Akhtar},
  title        = {Selective Neuron Amplification for Training-Free Task Enhancement},
  year         = {2026},
  howpublished = {arXiv preprint arXiv:2604.07098 [cs.LG]},
  note         = {Available: \url{https://arxiv.org/abs/2604.07098}}
}

\end{document}